%% file: main.tex
\documentclass[11pt,a4paper,dvipsnames]{article}
\usepackage[hyperref]{emnlp2020}
\usepackage{times}
\usepackage{latexsym}
\usepackage{todonotes}
\usepackage{booktabs}
\usepackage[utf8]{inputenc}
\usepackage{adjustbox}
\usepackage{multirow}
\usepackage{colortbl}
\usepackage{tikz}
\usepackage{xstring}
\usepackage{arydshln}
\usepackage{adjustbox}
\usepackage{array}
\usepackage{multicol}

\newcommand{\spc}{\texttt{\scriptsize \color{lightgray} \char32}}
\newcommand{\rspc}{\texttt{\scriptsize \color{Red} \char32}}

\makeatletter \def\adl@drawiv#1#2#3{%
        \hskip.5\tabcolsep
        \xleaders#3{#2.5\@tempdimb #1{1}#2.5\@tempdimb}%
                #2\z@ plus1fil minus1fil\relax
        \hskip.5\tabcolsep}
\newcommand{\cdashlinelr}[1]{%
  \noalign{\vskip\aboverulesep
           \global\let\@dashdrawstore\adl@draw
           \global\let\adl@draw\adl@drawiv}
  \cdashline{#1}
  \noalign{\global\let\adl@draw\@dashdrawstore
           \vskip\belowrulesep}}
\makeatother

\presetkeys%
    {todonotes}%
    {inline,backgroundcolor=yellow}{}

\newcounter{notecounter}
\newcommand{\enotesoff}{\long\gdef\enote##1##2{}}
\newcommand{\enoteson}{\long\gdef\enote##1##2{{
			\stepcounter{notecounter}
			\large\bf
			\hspace{100cm}\arabic{notecounter} $<<<$ ##1: ##2
			$>>>$\hspace{1cm}}}}
\enoteson
\enotesoff

\usepackage{microtype}

\aclfinalcopy 
\def\aclpaperid{1248} 

\title{Towards Reasonably-Sized Character-Level Transformer NMT \\ by
Finetuning Subword Systems}

\author{Jindřich Libovický \and Alexander Fraser \\
  Center for Information and Speech Processing \\
  Ludwig Maximilian University of Munich \\
  Munich, Germany \\
  \texttt{\{libovicky, fraser\}@cis.lmu.de}}

\date{}

\begin{document}
\maketitle
\begin{abstract}
    Applying the Transformer architecture on the character level usually
    requires very deep architectures that are difficult and slow to train.
%
    These problems can be partially overcome by incorporating a segmentation into tokens
    in the model.
    We show that by initially training a subword model
    and then finetuning it on characters, we can obtain a neural
    machine translation model that works at the character level without
    requiring token segmentation. We
%
    use only the
    vanilla 6-layer Transformer Base architecture.
%
    Our character-level models better capture morphological phenomena and show
    more
    robustness to
    noise at the expense of somewhat
    worse overall translation quality.
    Our study is a significant step towards high-performance and easy to train character-based
    models that are not extremely large.
\end{abstract}

\section{Introduction}\label{sec:introduction}

State-of-the-art neural machine translation (NMT) models operate almost
end-to-end except for input and output text segmentation. The segmentation is
done by first employing rule-based tokenization and then splitting into subword
units using statistical heuristics such as byte-pair encoding (BPE\@;
\citealp{sennrich-etal-2016-neural}) or SentencePiece
\citep{kudo-richardson-2018-sentencepiece}.

Recurrent sequence-to-sequence (S2S\@) models can learn translation end-to-end
(at the character level) without changes in the architecture
\citep{cherry-etal-2018-revisiting}, given sufficient model depth. Training
character-level Transformer S2S models \citep{vaswani2017attention} is more
complicated because the self-attention size is quadratic in the sequence
length.

In this paper, we empirically evaluate Transformer S2S models. We observe that
training a character-level model directly from random initialization suffers
from instabilities, often preventing it from converging.
Instead, we propose finetuning subword-based models to get a model without
explicit segmentation. Our character-level models show slightly worse
translation quality, but have better robustness towards input noise and better
capture morphological phenomena. Our approach is important
because previous approaches have relied on very large transformers, which are
out of reach for much of the research community.

\section{Related Work}\label{sec:related}

Character-level decoding seemed to be relatively easy with recurrent S2S models
\citep{chung-etal-2016-character}. But early attempts at achieving
segmentation-free NMT with recurrent networks used input hidden states covering
a constant character span \citep{lee-etal-2017-fully}.
\citet{cherry-etal-2018-revisiting} showed that with a sufficiently deep
recurrent model, no changes in the model are necessary, and they can still
reach translation quality that is on par with subword models.
\citet{luong-manning-2016-achieving} and \citet{ataman-etal-2019-importance}
can leverage character-level information but they require tokenized text as an
input and only have access to the character-level embeddings of predefined
tokens.

Training character-level transformers is more challenging.
\citet{choe2019bridging} successfully trained a character-level left-to-right
Transformer language model that performs on par with a subword-level model.
However, they needed a large model with 40 layers trained on a billion-word
corpus, with prohibitive computational costs.

In the most related work to ours, \citet{gupta2019characterbased} managed to
train a character-level NMT with the Transformer model using Transparent
Attention \citep{bapna-etal-2018-training}. Transparent attention attends to
all encoder layers simultaneously, making the model more densely connected but
also more computationally expensive. During training, this improves the
gradient flow from the decoder to the encoder. \citet{gupta2019characterbased}
claim that Transparent Attention is crucial for training character-level
models, and show results on very deep networks, with similar results in terms
of translation quality and model robustness to ours.  In contrast, our model,
which is not very deep, trains quickly. It also supports fast inference and
uses less RAM, both of which are important for deployment.

\citet{gao-etal-2020-character} recently proposed adding a convolutional
sub-layer in the Transformer layers. At the cost of a 30\% increase of model
parameter count, they managed to narrow the gap between subword- and
character-based models to half. Similar results were also reported by
\citet{banar2020character}, who reused the convolutional preprocessing layer
with constant step segments \citet{lee-etal-2017-fully} in a Transformer model.

\section{Our Method}\label{sec:method}

We train our character-level models by finetuning subword models, which does
not increase the number of model parameters. Similar to the transfer learning
experiments of \citet{kocmi-bojar-2018-trivial}, we start with a fully trained
subword model and continue training with the same data segmented using only a
subset of the original vocabulary.

To stop the initial subword models from relying on sophisticated tokenization
rules, we opt for the loss-less tokenization algorithm from SentencePiece
\citep{kudo-richardson-2018-sentencepiece}. First, we replace all spaces with
the \texttt{\small\_} sign and do splits before all non-alphanumerical
characters (first line of Table~\ref{tab:segmentation}).  In further
segmentation, the special space sign \texttt{\small\_} is treated identically
to other characters.

\begin{table}

    \centering
    \scalebox{1.0}{%
    \begin{tabular}{cp{.75\columnwidth}}
    \toprule

    \multirow{2}{*}{\parbox{0.9cm}{\centering tokeni\-zation}}
    & The\spc{}cat\spc{}sleeps\spc{}on\spc{}a\spc{}mat. \\
    &\_The\spc\_cat\spc\_sleeps\spc\_on\spc\_a\spc\_mat\spc. \\

         \midrule

         32k  &  \_The\spc\_cat\spc{}\_sle\spc{}eps\spc\_on\spc{}\_a\spc\_mat\spc. \\

         8k   &  \_The\spc\_c\spc{}at\spc{}\_s\spc{}le\spc{}eps\spc\_on\spc{}\_a\spc\_m\spc{}at\spc{}. \\

         500  &  \_The\spc\_c\spc{}at\spc{}\_s\spc{}le\spc{}ep\spc{}s\spc\_on\spc{}\_a\spc\_m\spc{}at\spc. \\

         0    &  \_\rspc{}T\spc{}h\spc{}e\spc\_\rspc{}c\spc{}a\spc{}t\spc\_\rspc{}s\spc{}l\spc{}e\spc{}e\spc{}p\spc{}s\spc\_\rspc{}o\spc{}n\spc\_\rspc{}\\ & a\spc\_\rspc{}m\spc{}a\spc{}t\spc.  \\

    \bottomrule

    \end{tabular}}

    \caption{Examples of text tokenization and subword segmentation with
      different
    numbers of BPE merges.}\label{tab:segmentation}
\end{table}

\begin{table}[t]

    \centering \scalebox{1.0}{\input{data_stats}}

    \caption{Statistics of English-German parallel data under different
    segmentations.}\label{tab:segmentation_stats}

\end{table}

We use BPE~\cite{sennrich-etal-2016-neural} for subword segmentation because it
generates the merge operations in a deterministic order. Therefore, a
vocabulary based on a smaller number of merges is a subset of vocabulary based
on more merges estimated from the same training data.
Examples of the segmentation are provided in Table~\ref{tab:segmentation}.
Quantitative effects of different segmentation on the data are presented in
Table~\ref{tab:segmentation_stats}, showing that character sequences are on
average more than 4 times longer than subword sequences with 32k vocabulary.

We experiment both with deterministic segmentation and stochastic segmentation
using BPE Dropout \citep{provilkov-etal-2020-bpe}. At training time, BPE
Dropout randomly discards BPE merges with probability $p$, a hyperparameter of
the method.  As a result of this, the text gets stochastically segmented into
smaller units. BPE Dropout increases translation robustness on the source side
but typically has a negative effect when used on the target side. In our
experiments, we use BPE Dropout both on the source and target side. In this
way, the character-segmented inputs will appear already at training time,
making the transfer learning easier.

We test two methods for finetuning subword models to reach character-level
models: first, direct finetuning of subword models, and second, iteratively
removing BPE merges in several steps in a curriculum learning setup
\citep{bengio2009curriculum}. In both cases we always finetune models until
they are fully converged, using early stopping.

\begin{table*}[t!]

    \input{table_less_numbers}

    \caption{Quantitative results of the experiments with deterministic
    segmentation. The left part of the table shows subword-based models trained
    from random initialization, the right part shows character-level models
    trained by finetuning. The yellower the background color, the better the
    value. Small numbers denote the difference from the best model, $\ast$ is
    the best model. For finetuning experiments (on the right) we report both
    difference from the best model and from the parent model. Validation BLEU
    score are in in the Appendix.}\label{tab:results}

\end{table*}

\section{Experiments}\label{sec:experiments}

To cover target languages of various morphological complexity, we conduct our
main experiments on two resource-rich language pairs, English-German and
English-Czech; and on a low-resource pair, English-Turkish.  Rich inflection in
Czech, compounding in German, and agglutination in Turkish are examples of
interesting phenomena for character models.
We train and evaluate the English-German translation using the 4.5M parallel
sentences of the WMT14 data \citep{bojar-etal-2014-findings}. Czech-English is
trained on 15.8M sentence pairs of the CzEng 1.7 corpus \citep{bojar2016czeng}
and tested on WMT18 data \citep{bojar-etal-2018-findings}. English-to-Turkish
translation is trained on 207k sentences of the SETIMES2 corpus
\citep{tiedemann-2012-parallel} and evaluated on the WMT18 test set.

We follow the original hyperparameters for the Transformer Base model
\citep{vaswani2017attention}, including the learning rate schedule. For
finetuning, we use Adam \citep{kingma2015adam} with a constant learning rate
$10^{-5}$. All models are trained using Marian
\citep{junczys-dowmunt-etal-2018-marian-fast}.  We also present results for
character-level English-German models having about the same number of
parameters as the best-performing subword models. In experiments with BPE
Dropout, we set dropout probability $p=0.1$.

We evaluate the translation quality using BLEU \citep{papineni-etal-2002-bleu},
chrF \citep{popovic-2015-chrf}, and METEOR 1.5
\citep{denkowski-lavie-2014-meteor}.
Following \citet{gupta2019characterbased}, we also conduct a noise-sensitivity
evaluation to natural noise as introduced by \citet{belinkov2018syntheic}. With
probability
$p$
words are replaced with their
variants from a misspelling corpus.
Following \citet{gupta2019characterbased}, we assume the BLEU scores measured
with input can be explained by a linear approximation with intercept $\alpha$
and slope $\beta$ using the noise probability $p$:
$ \mathrm{BLEU} \approx \beta p + \alpha. $
However, unlike them, we report the relative translation quality degradation
$\beta / \alpha$ instead of only $\beta$.
Parameter $\beta$ corresponds to
absolute BLEU score degradation and is thus higher given lower-quality systems,
making them seemingly more robust.



To look at morphological generalization, we evaluate translation into Czech and
German using MorphEval \citep{burlot-yvon-2017-evaluating}. MorphEval consists
of 13k sentence pairs that differ in exactly one morphological category. The
score is the percentage of pairs where the correct sentence is preferred.


\section{Results}\label{sec:results}

The results of the experiments are presented in Table~\ref{tab:results}.
The translation quality only slightly decreases when drastically decreasing the
vocabulary. However, there is a gap between the character-level and
subword-level model of 1--2 BLEU points. With the exception of Turkish, models
trained by finetuning reach by a large margin better translation quality than
character-level models trained from scratch.

In accordance with \citet{provilkov-etal-2020-bpe}, we found that BPE Dropout
applied both on the source and target side leads to slightly worse translation
quality, presumably because the stochastic segmentation leads to multimodal
target distributions. The detailed results are presented in
Appendix~\ref{sec:appendixdropout}. However, for most language pairs, we found
a small positive effect of BPE dropout on the finetuned systems (see
Table~\ref{tab:bpedropout}).

\begin{table}

    \centering
    \begin{tabular}{lcccc}
        \toprule

        \multirow{2}{*}{Direction}  &
        \multicolumn{2}{c}{Determ. BPE} &
			\multicolumn{2}{c}{BPE Dropout}  \\ \cmidrule(rl){2-3} \cmidrule(rl){4-5}

        & BLEU & chrF & BLEU & chrF \\ \midrule

        en-de & 25.2 & .559  & 24.9 & .560 \\

        de-en & 28.2 & .562  & 28.5 & .564 \\

        en-cs & 19.3 & .447  & 19.5 & .480 \\

        en-tr & 12.0 & .456  & 12.3 & .460 \\

        \bottomrule
    \end{tabular}

    \caption{BLEU scores of character-level models trained by finetuning of the
    systems with 500 token vocabularies using deterministic BPE segmetnation
    and BPE dropout.}\label{tab:bpedropout}

\end{table}

For English-to-Czech translation, we observe a large drop in BLEU score with
the decreasing vocabulary size, but almost no drop in terms of METEOR score,
whereas for other language pairs, all metrics are in agreement.
The differences between the subword and character-level models are less
pronounced in the low-resourced English-to-Turkish translation.
%

\begin{table}

    \small\centering
    \begin{tabular}{llcc}
        \toprule
        vocab. & architecture & \# param. & BLEU \\ \midrule

        BPE 16k & Base                &   42.6M   & 26.86 \\
        char.\  & Base                &   35.2M   & 25.21 \\
        char.\  & Base + FF dim. 2650 &   42.6M   & 25.37 \\

        \bottomrule
    \end{tabular}

    \caption{Effect of model size on translation quality for Engslih-to-German
    translation.}\label{tab:parameters}

\end{table}

Whereas the number of parameters in transformer layers in all models is
constant at 35 million, the number of parameters in the embeddings decreases
30$\times$ from over 15M to only slightly over 0.5M, with overall a 30\%
parameter count reduction. However, matching the number of parameters by
increasing the model capacity narrows close the performance gap, as shown in
Table~\ref{tab:parameters}.

In our first set of experiments, we finetuned the model using the
character-level input directly. Experiments with parent models of various
vocabulary sizes (column ``Direct finetuning'' in Table~\ref{tab:results})
suggest the larger the parent vocabulary, the worse the character-level
translation quality. This results led us to hypothesize that gradually
decreasing the vocabulary size in several steps might lead to better
translation quality.  In the follow-up experiment, we gradually reduced the
vocabulary size by 500 and always finetuned until convergence. But we observed
a small drop in translation quality in every step, and the overall translation
quality was slightly worse than with direct finetuning (column ``In steps'' in
Table~\ref{tab:results}).

\begin{figure}
    %
    \includegraphics{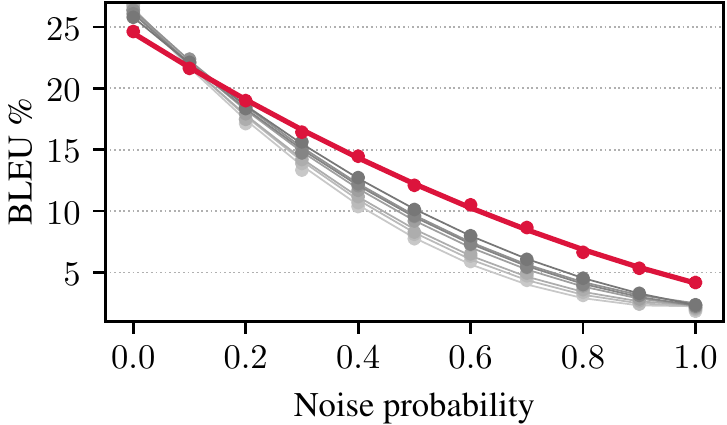}
    \caption{Degradation of the translation quality of the subword (gray, the
    darker the color, the smaller vocabulary) and character-based systems (red)
    for English-German translation with increasing noise.}\label{fig:noise}

\end{figure}

With our character-level models, we achieved higher robustness towards
source-side noise (Figure~\ref{fig:noise}). Models trained with a smaller
vocabulary tend to be more robust towards source-side noise.

Character-level models tend to perform slightly better in the MorphEval
benchmark. Detailed results are shown in Table~\ref{tab:morpheval}. In German,
this is due to better capturing of agreement in coordination and future tense.
This result is unexpected because these phenomena involve long-distance
dependencies.  On the other hand, the character-level models perform worse on
compounds, which are a local phenomenon. \citet{ataman-etal-2019-importance}
observed similar results on compounds in their hybrid character-word-level
method. We suspect this might be caused by poor memorization of some compounds
in the character models.

\begin{table}[t]

    \centering\small
	\input{morpheval}

    \caption{MorphEval Results for English to
    German and English to Czech.}\label{tab:morpheval}

\end{table}

In Czech, models with a smaller vocabulary better cover agreement in gender and
number in pronouns, probably due to direct access to inflective endings.
Unlike German, character-level models capture worse agreement in coordinations,
presumably due to there being a longer distance in characters.

\begin{table}[t]
    \centering
    \scalebox{0.76}{%
    \begin{tabular}{ccccccccc}
\toprule
 & 32k & 16k & 8k & 4k & 2k & 1k & 500 & 0 \\ \midrule

T & 1297 & 1378 & 1331 & 1151 & 1048 & 903 & 776 & 242 \\

I & 21.8 & 18.3 & 17.2 & 12.3 & 12.3 & 8.8 & 7.3 & 3.9 \\ \midrule

B & 26.9 & 26.9 & 26.7 & 26.4 & 26.4 & 26.1 & 25.8 & 25.2 \\

\bottomrule
    \end{tabular}}

    \caption{Training (T) and inference (I) speed in sentences processed per
    second on a single GPU (GeForce GTX 1080 Ti) compared to BLEU scores (B)
    for English-German translation.}\label{tab:times}

\end{table}

Training and inference times are shown in Table~\ref{tab:times}. Significantly
longer sequences also manifest in slower training and inference.
Table~\ref{tab:times} shows that our character-level models are 5--6$\times$
slower than subword models with 32k units. Doubling the number of layers, which
had a similar effect on translation quality as the proposed finetuning
\citep{gupta2019characterbased}, increases the inference time approximately
2--3$\times$ in our setup.

\section{Conclusions}\label{sec:conclusions}

We presented a simple approach for training character-level models by
finetuning subword models. Our approach does not require computationally
expensive architecture changes and does not require dramatically increased
model depth.
Subword-based models can be finetuned to work on the character level without
explicit segmentation with somewhat of a drop in translation quality.  The
models are robust to input noise and better capture some morphological
phenomena.  This is important for research groups that need to train and deploy
character Transformer models without access to very large computational
resources.

\enote{AF}{NOT DONE: I usually do not include future work in a paper submission. But here
I'd be tempted to say that this is an important step to switching to
character-level models for everything (and maybe also to reiterate that this is
highly desirable). And, if possible, also to speculate on what we will have to do to get
superior performance to (or at least the same performance as) subword models.}

\enote{AF}{I'd delete ``Association for Computational Linguistics'', the
publisher entry, from the bibtex entries.}

\section*{Acknowledgments}

The work was supported by the European Research Council (ERC) under the
European Union’s Horizon 2020 research and innovation programme (grant
agreement No.~640550) and by German Research Foundation (DFG; grant FR
2829/4-1).

\bibliography{anthology,references}
\bibliographystyle{acl_natbib}

\appendix

\begin{table*}[t!]

\input{bpe_dropout_table}

	\caption{Comparison of the trasnaltion quality without (gray numbers) and
		with BPE Dropout (with the same color coding as in
		Table~\ref{tab:results}).}\label{tab:bpedropoutBig}

\end{table*}

\begin{table*}[b]

    \input{validation_table}

    \caption{BLEU scores on the validation data: WMT13 test set for
    English-German in both directions, WMT17 test set for English-Czech and
    English-Turkish directions.}\label{tab:validation}

\end{table*}

\section{Effect of BPE Dropout}\label{sec:appendixdropout}

We discussed the effect of BPE dropout in Section~\ref{tab:results}.
Table~\ref{tab:bpedropoutBig} shows the comparison of the main quantitative
results with and without BPE dropout.

\section{Notes on Reproducibility}

The training times were measured on machines with GeForce GTX 1080 Ti GPUs and
with Intel Xeon E5--2630v4 CPUs (2.20GHz). The parent models were trained on 4
GPUs simultaneously, the finetuning experiments were done on a single GPU\@.

We used model hyperparameters used by previous work and did not experiment with
the hyperparameters of the architecture and training of the initial models.
The only hyperparameter that we tuned was the learning rate of the finetuning.
We set the value to $10^{-5}$ after several experiments with English-to-German
translation with values between $10^{-7}$ and $10^{-3}$ based on the BLEU score
on validation data.

We downloaded the training data from the official WMT web
(\url{http://www.statmt.org/wmt18/}).The test and validation sets were
downloaded using SacreBleu (\url{https://github.com/mjpost/sacreBLEU}). The BPE
segmentation is done using FastBPE (\url{https://github.com/glample/fastBPE}).
For BPE Dropout, we used YouTokenToMe
(\url{https://github.com/VKCOM/YouTokenToMe}). A script that downloads and
pre-processes the data is attached to the source code. It also includes
generating the noisy synthetic data (using
\url{https://github.com/ybisk/charNMT-noise}) and preparing data and tools
required by MorphEval (\url{https://github.com/franckbrl/morpheval}).

The models were trained and evaluated with Marian v1.7.0
(\url{https://github.com/marian-nmt/marian/releases/tag/1.7.0}).

Validation BLEU scores are tabulated in Table~\ref{tab:validation}.

\end{document}

%% file: data_stats.tex
\begin{tabular}{cccccc}

\toprule
\# merges &
\multirow{2}{1cm}{\centering segm.~/ sent.} &
\multirow{2}{1cm}{\centering segm.~/ token} &
\multicolumn{2}{c}{\centering avg. unit size}
  \\ \cmidrule(lr){4-5}
& & & en & de  \\ \midrule

32k &  28.4 & 1.3 & 4.37 & 4.51 \\
16k &  31.8 & 1.4 & 3.95 & 3.98 \\
8k  &  36.2 & 1.6 & 3.46 & 3.50 \\
4k  &  41.5 & 1.9 & 3.03 & 3.04 \\
2k  &  47.4 & 2.1 & 2.66 & 2.67 \\
1k  &  54.0 & 2.4 & 2.32 & 2.36 \\
500 &  61.4 & 2.7 & 2.03 & 2.08 \\ \midrule
0   & 126.1 & 5.6 & 1.00 & 1.00 \\ \bottomrule

\end{tabular}

%% file: table_less_numbers.tex
\newcommand\cellScaleColor[2]
  {\colorbox{GreenYellow!#1}{#2}}

\newcommand{\Bende}[1]{%
\pgfmathsetmacro{\Min}{22.57}%
\pgfmathsetmacro{\Max}{26.86}%
\pgfmathsetmacro{\PercentColor}{max(min(100.0 * (#1 - \Min)/(\Max - \Min), 100.0), 0.00)}%
\cellScaleColor{\PercentColor}{#1}
}
\newcommand{\Cende}[1]{%
\pgfmathsetmacro{\Min}{0.5262}%
\pgfmathsetmacro{\Max}{0.5692}%
\pgfmathsetmacro{\PercentColor}{max(min(100.0 * (#1 - \Min)/(\Max - \Min), 100.0), 0.00)}%
\cellScaleColor{\PercentColor}{#1}
}
\newcommand{\Mende}[1]{%
\pgfmathsetmacro{\Min}{45.00}%
\pgfmathsetmacro{\Max}{47.95}%
\pgfmathsetmacro{\PercentColor}{max(min(100.0 * (#1 - \Min)/(\Max - \Min), 100.0), 0.00)}%
\cellScaleColor{\PercentColor}{#1}
}
\newcommand{\Nende}[1]{%
\pgfmathsetmacro{\Min}{-1.075}%
\pgfmathsetmacro{\Max}{-0.990}%
\pgfmathsetmacro{\PercentColor}{max(min(100.0 * (#1 - \Min)/(\Max - \Min), 100.0), 0.00)}%
\cellScaleColor{\PercentColor}{#1}
}

\newcommand{\Rende}[1]{%
\pgfmathsetmacro{\Min}{89.18}%
\pgfmathsetmacro{\Max}{90.10}%
\pgfmathsetmacro{\PercentColor}{max(min(100.0 * (#1 - \Min)/(\Max - \Min), 100.0), 0.00)}%
\cellScaleColor{\PercentColor}{#1}
}

\newcommand{\Bdeen}[1]{%
\pgfmathsetmacro{\Min}{26.62}%
\pgfmathsetmacro{\Max}{30.14}%
\pgfmathsetmacro{\PercentColor}{max(min(100.0 * (#1 - \Min)/(\Max - \Min), 100.0), 0.00)}%
\cellScaleColor{\PercentColor}{#1}
}

\newcommand{\Cdeen}[1]{%
\pgfmathsetmacro{\Min}{.5433}%
\pgfmathsetmacro{\Max}{.5732}%
\pgfmathsetmacro{\PercentColor}{max(min(100.0 * (#1 - \Min)/(\Max - \Min), 100.0), 0.00)}%
\cellScaleColor{\PercentColor}{#1}
}

\newcommand{\Mdeen}[1]{%
\pgfmathsetmacro{\Min}{35.07}%
\pgfmathsetmacro{\Max}{37.42}%
\pgfmathsetmacro{\PercentColor}{max(min(100.0 * (#1 - \Min)/(\Max - \Min), 100.0), 0.00)}%
\cellScaleColor{\PercentColor}{#1}
}

\newcommand{\Ndeen}[1]{%
\pgfmathsetmacro{\Min}{-0.451}%
\pgfmathsetmacro{\Max}{-0.364}%
\pgfmathsetmacro{\PercentColor}{max(min(100.0 * (#1 - \Min)/(\Max - \Min), 100.0), 0.00)}%
\cellScaleColor{\PercentColor}{#1}
}

\newcommand{\Bencs}[1]{%
\pgfmathsetmacro{\Min}{18.24}%
\pgfmathsetmacro{\Max}{21.07}%
\pgfmathsetmacro{\PercentColor}{max(min(100.0 * (#1 - \Min)/(\Max - \Min), 100.0), 0.00)}%
\cellScaleColor{\PercentColor}{#1}
}

\newcommand{\Cencs}[1]{%
\pgfmathsetmacro{\Min}{0.4645}%
\pgfmathsetmacro{\Max}{0.4899}%
\pgfmathsetmacro{\PercentColor}{max(min(100.0 * (#1 - \Min)/(\Max - \Min), 100.0), 0.00)}%
\cellScaleColor{\PercentColor}{#1}
}

\newcommand{\Mencs}[1]{%
\pgfmathsetmacro{\Min}{24.57}%
\pgfmathsetmacro{\Max}{25.98}%
\pgfmathsetmacro{\PercentColor}{max(min(100.0 * (#1 - \Min)/(\Max - \Min), 100.0), 0.00)}%
\cellScaleColor{\PercentColor}{#1}
}

\newcommand{\Nencs}[1]{%
\pgfmathsetmacro{\Min}{-1.032}%
\pgfmathsetmacro{\Max}{-0.816}%
\pgfmathsetmacro{\PercentColor}{max(min(100.0 * (#1 - \Min)/(\Max - \Min), 100.0), 0.00)}%
\cellScaleColor{\PercentColor}{#1}
}

\newcommand{\Rencs}[1]{%
\pgfmathsetmacro{\Min}{81.28}%
\pgfmathsetmacro{\Max}{84.87}%
\pgfmathsetmacro{\PercentColor}{max(min(100.0 * (#1 - \Min)/(\Max - \Min), 100.0), 0.00)}%
\cellScaleColor{\PercentColor}{#1}
}

\newcommand{\Tencs}[1]{%
\pgfmathsetmacro{\Min}{4.48}%
\pgfmathsetmacro{\Max}{5.84}%
\pgfmathsetmacro{\PercentColor}{max(min(100.0 * (#1 - \Min)/(\Max - \Min), 100.0), 0.00)}%
\cellScaleColor{\PercentColor}{#1}
}

\newcommand{\Bentr}[1]{%
\pgfmathsetmacro{\Min}{11.55}%
\pgfmathsetmacro{\Max}{13.11}%
\pgfmathsetmacro{\PercentColor}{max(min(100.0 * (#1 - \Min)/(\Max - \Min), 100.0), 0.00)}%
\cellScaleColor{\PercentColor}{#1}
}

\newcommand{\Centr}[1]{%
\pgfmathsetmacro{\Min}{0.4500}%
\pgfmathsetmacro{\Max}{0.4620}%
\pgfmathsetmacro{\PercentColor}{max(min(100.0 * (#1 - \Min)/(\Max - \Min), 100.0), 0.00)}%
\cellScaleColor{\PercentColor}{#1}
}

\newcommand{\Nentr}[1]{%
\pgfmathsetmacro{\Min}{-0.916}%
\pgfmathsetmacro{\Max}{-0.657}%
\pgfmathsetmacro{\PercentColor}{max(min(100.0 * (#1 - \Min)/(\Max - \Min), 100.0), 0.00)}%
\cellScaleColor{\PercentColor}{#1}
}

\newcommand\compare[2]{%
    \scriptsize \IfSubStr{#1}{-}{\color{BrickRed} #1}{\color{ForestGreen} #1} \color{black} {\tiny /} \IfSubStr{#2}{-}{\color{BrickRed} #2}{\color{ForestGreen} #2}
}
\newcommand\fromBest[1]{\scriptsize \textcolor{BrickRed}{#1}}
\newcommand\best{\scriptsize $\ast$}

\newcolumntype{C}{>{\centering\arraybackslash}p{1.08cm}}

\adjustbox{max width=\textwidth}{%
\begin{tabular}{ll|CCCCCCCC|ccc|c}
\toprule

& & \multicolumn{8}{c|}{From random initialization}
& \multicolumn{3}{c|}{Direct finetuning from}
& \multirow{2}{*}{In steps} \\

\cmidrule(lr){3-10} \cmidrule(lr){11-13} 

& & 32k & 16k & 8k & 4k & 2k & 1k & 500 & 0 &
500 & 1k & 2k & 
\\ \midrule


\multirow{6}{*}{\rotatebox{90}{en-de}}
& \multirow{2}{*}{BLEU} &
    \Bende{26.9} & \Bende{26.9} & \Bende{26.7} & \Bende{26.4} & \Bende{26.4} & \Bende{26.1} & \Bende{25.8} & \Bende{22.6} &
    \Bende{25.2} & \Bende{25.0} & \Bende{25.0} & 
    \Bende{24.6}

\\

   &  &  
    \fromBest{-0.03} & \best & \fromBest{-0.20} & \fromBest{-0.47} & \fromBest{-0.50} & \fromBest{-0.78} & \fromBest{-1.07} & \fromBest{-4.29} &
    \compare{-1.65}{-0.58} & \compare{-1.88}{-1.10} & \compare{-1.85}{-0.78} &
    \compare{-2.23}{-1.16}
\\
& chrF &
    \Cende{.569} & \Cende{.568} & \Cende{.568} & \Cende{.568} & \Cende{.564} & \Cende{.564} & \Cende{.561} & \Cende{.526} &
    \Cende{.559} & \Cende{.559} & \Cende{.559} & 
    \Cende{.556}
\\

%

& METEOR &
    \Mende{47.7} & \Mende{48.0} & \Mende{47.9} & \Mende{47.8} & \Mende{47.9} & \Mende{47.7} & \Mende{47.6} & \Mende{45.0} &
    \Mende{46.5} & \Mende{46.4} & \Mende{46.4} & 
    \Mende{46.3}

\\ \cdashlinelr{2-14}

& Noise sens. &
    \Nende{-1.07} & \Nende{-1.06} & \Nende{-1.05} & \Nende{-1.03} & \Nende{-1.01} & \Nende{-1.02} & \Nende{-1.00} & \Nende{-0.85} &
    \Nende{-0.99} & \Nende{-0.99} & \Nende{-0.99} &
    \Nende{-0.88}
\\

& MorphEval &
    \Rende{90.0} & \Rende{89.5} & \Rende{89.4} &  \Rende{89.6} &  \Rende{89.8} &  \Rende{90.0} &  \Rende{89.2} & \Rende{89.2} &
    \Rende{89.9} & \Rende{90.3} & \Rende{89.3} &
    \Rende{90.1}

\\ \midrule


\multirow{5}{*}{\rotatebox{90}{de-en}}
& \multirow{2}{*}{BLEU} &
    \Bdeen{29.8} & \Bdeen{30.1} & \Bdeen{29.6} & \Bdeen{29.3} & \Bdeen{28.6} & \Bdeen{28.5} & \Bdeen{28.1} & \Bdeen{26.6} &
    \Bdeen{28.2} & \Bdeen{28.4} & \Bdeen{27.7} & 
    \Bdeen{28.2}
\\

   &  &  
   \fromBest{-0.34} & \best & \fromBest{-0.53} & \fromBest{-0.83} & \fromBest{-1.62} & \fromBest{-1.67} & \fromBest{-1.99} & \fromBest{-3.51} &
    \compare{-1.94}{+0.05} & \compare{-1.76}{-0.10} & \compare{-2.52}{-0.90} &
    \compare{-1.89}{+0.10}
\\
& chrF &
    \Cdeen{.570} & \Cdeen{.573} & \Cdeen{.568} & \Cdeen{.567} & \Cdeen{.562} & \Cdeen{.558} & \Cdeen{.558} & \Cdeen{.543} &
    \Cdeen{.562} & \Cdeen{.564} & \Cdeen{.559} & 
    \Cdeen{.563}
\\
& METEOR &
    \Mdeen{37.1} & \Mdeen{37.4} & \Mdeen{37.2} & \Mdeen{37.2} & \Mdeen{36.9} & \Mdeen{37.2} & \Mdeen{36.9} & \Mdeen{35.1} &
    \Mdeen{36.4} & \Mdeen{36.4} & \Mdeen{36.0} & 
    \Mdeen{36.4}

\\ \cdashlinelr{2-14}

& Noise sens.&
    \Ndeen{-0.45} & \Ndeen{-0.43} & \Ndeen{-0.41} & \Ndeen{-0.42} & \Ndeen{-0.43} & \Ndeen{-0.42} & \Ndeen{-0.41} & \Ndeen{-0.30} &
    \Ndeen{-0.37} & \Ndeen{-0.37} & \Ndeen{-0.37} &
    \Ndeen{-0.36}
\\ \midrule


\multirow{8}{*}{\rotatebox{90}{en-cs}}
& BLEU &
    \Bencs{21.1} & \Bencs{20.8} & \Bencs{20.9} & \Bencs{20.6} & \Bencs{20.1} & \Bencs{20.0} & \Bencs{19.5} & \Bencs{18.2} &
    \Bencs{19.2} & \Bencs{19.3} & \Bencs{19.4} & 
    \Bencs{19.3}
\\
   &  &  
   \best{} & \fromBest{-0.25} & \fromBest{-0.13} & \fromBest{-0.46} & \fromBest{-0.96} & \fromBest{-1.05} & \fromBest{-1.54} & \fromBest{-2.82} &
    \compare{-1.81}{-0.27} & \compare{-1.73}{-0.68} & \compare{-1.64}{-0.68} &
    \compare{-1.81}{-0.27}
\\
& chrF &
    \Cencs{.489} & \Cencs{.488} & \Cencs{.490} & \Cencs{.487} & \Cencs{.483} & \Cencs{.482} & \Cencs{.478} & \Cencs{.465} &
    \Cencs{.477} & \Cencs{.476} & \Cencs{.478} & 
    \Cencs{.477}
\\
& METEOR &
    \Mencs{26.0} & \Mencs{25.8} & \Mencs{26.0} & \Mencs{25.8} & \Mencs{25.7} & \Mencs{25.7} & \Mencs{25.4} & \Mencs{24.6} &
    \Mencs{25.2} & \Mencs{25.2} & \Mencs{25.2} & 
    \Mencs{25.1}
\\ \cdashlinelr{2-14}
& Noise sens. &
    \Nencs{-1.03} & \Nencs{-1.01} & \Nencs{-1.01} & \Nencs{-1.01} & \Nencs{-0.94} & \Nencs{-0.93} & \Nencs{-0.91} & \Nencs{-0.79} &
    \Nencs{-0.82} & \Nencs{-0.84} & \Nencs{-0.87} &
    \Nencs{-0.82}
\\
& MorphEval &
    \Rencs{83.9} & \Rencs{84.6} & \Rencs{83.7} & \Rencs{83.9} & \Rencs{84.3} & \Rencs{84.4} & \Rencs{84.7} & \Rencs{82.1} &
    \Rencs{84.7} & \Rencs{84.1} & \Rencs{81.9} &
    \Rencs{81.3}
\\ \midrule


\multirow{4}{*}{\rotatebox{90}{en-tr}}
& BLEU &
    \Bentr{12.6} & \Bentr{13.1} & \Bentr{12.7} & \Bentr{12.8} & \Bentr{12.5} & \Bentr{12.3} & \Bentr{12.2} & \Bentr{12.4} &
    \Bentr{12.0} & \Bentr{12.6} & \Bentr{12.3} &
    \Bentr{11.6}
\\
   &  &  
    \fromBest{-0.48} & \best{} & \fromBest{-0.36} & \fromBest{-0.29} & \fromBest{-0.58} & \fromBest{-0.77} & \fromBest{-0.86} & \fromBest{-.0.73} &
    \compare{-1.08}{-0.22} & \compare{-0.85}{-0.08} & \compare{-0.82}{-0.53} &
    \compare{-1.54}{-0.68}
\\
& chrF &
    \Centr{.455} & \Centr{.462} & \Centr{.459} & \Centr{.456} & \Centr{.457} & \Centr{.457} & \Centr{.455} & \Centr{.461} &
    \Centr{.456} & \Centr{.460} & \Centr{.459} &
    \Centr{.450}
\\ \cdashlinelr{2-14}

& Noise sens. &
    \Nentr{-0.99} & \Nentr{-0.91} & \Nentr{-0.90} & \Nentr{-0.87} & \Nentr{-0.85} & \Nentr{-0.83} & \Nentr{-0.79} & \Nentr{-0.62} &
    \Nentr{-0.66} & \Nentr{-0.66} & \Nentr{-0.66} &
    \Nentr{-0.68}

\\ \bottomrule

\end{tabular}}

%% file: morpheval.tex
\begin{tabular}{lcccc}
\toprule
 & \multicolumn{2}{c}{en-de}  & \multicolumn{2}{c}{en-cs} \\ \cmidrule(lr){2-3} \cmidrule(lr){4-5}
 & BPE16k & char & BPE16k & char \\ \midrule

Adj.\ strong        & 95.5 & 97.2 & ---  & --- \\
Comparative         & 93.4 & 91.5 & 78.0 & 78.2 \\
Compounds           & 63.6 & 60.4 & ---  & --- \\
Conditional         & 92.7 & 92.3 & 45.8 & 47.6 \\
Coordverb-number    & 96.2 & 98.1 & 83.0 & 78.8 \\
Coordverb-person    & 96.4 & 98.1 & 83.2 & 78.6 \\
Coordverb-tense     & 96.6 & 97.8 & 79.2 & 74.8 \\
Coref.\ gender      & 94.8 & 92.8 & 74.0 & 75.8 \\
Future              & 82.1 & 89.0 & 84.4 & 83.8 \\
Negation            & 98.8 & 98.4 & 96.2 & 98.0 \\
Noun Number         & 65.5 & 66.6 & 78.6 & 79.2 \\
Past                & 89.9 & 90.1 & 88.8 & 87.4 \\
Prepositions        & ---  & ---  & 91.7 & 94.1 \\
Pronoun gender      & ---  & ---  & 92.6 & 92.2 \\
Pronoun plural      & 98.4 & 98.8 & 90.4 & 92.8 \\
Rel.\ pron.\ gender & 71.3 & 71.3 & 74.8 & 80.1 \\
Rel.\ pron.\ number & 71.3 & 71.3 & 76.6 & 80.9 \\
Superlative         & 98.9 & 99.8 & 92.0 & 92.0 \\
Verb position       & 95.4 & 94.2 & ---  & --- \\
\bottomrule
\end{tabular}

%% file: bpe_dropout_table.tex
\newcommand\cellScaleColor[2]
  {\colorbox{GreenYellow!#1}{#2}}

\newcommand{\Bende}[1]{%
\pgfmathsetmacro{\Min}{22.5}%
\pgfmathsetmacro{\Max}{26.86}%
\pgfmathsetmacro{\PercentColor}{max(min(100.0 * (#1 - \Min)/(\Max - \Min), 100.0), 0.00)}%
\cellScaleColor{\PercentColor}{#1}
}
\newcommand{\Cende}[1]{%
\pgfmathsetmacro{\Min}{0.526}%
\pgfmathsetmacro{\Max}{0.5692}%
\pgfmathsetmacro{\PercentColor}{max(min(100.0 * (#1 - \Min)/(\Max - \Min), 100.0), 0.00)}%
\cellScaleColor{\PercentColor}{#1}
}
\newcommand{\Mende}[1]{%
\pgfmathsetmacro{\Min}{44.90}%
\pgfmathsetmacro{\Max}{47.95}%
\pgfmathsetmacro{\PercentColor}{max(min(100.0 * (#1 - \Min)/(\Max - \Min), 100.0), 0.00)}%
\cellScaleColor{\PercentColor}{#1}
}
\newcommand{\Nende}[1]{%
\pgfmathsetmacro{\Min}{-1.075}%
\pgfmathsetmacro{\Max}{-0.990}%
\pgfmathsetmacro{\PercentColor}{max(min(100.0 * (#1 - \Min)/(\Max - \Min), 100.0), 0.00)}%
\cellScaleColor{\PercentColor}{#1}
}

\newcommand{\Rende}[1]{%
\pgfmathsetmacro{\Min}{89.18}%
\pgfmathsetmacro{\Max}{90.10}%
\pgfmathsetmacro{\PercentColor}{max(min(100.0 * (#1 - \Min)/(\Max - \Min), 100.0), 0.00)}%
\cellScaleColor{\PercentColor}{#1}
}

\newcommand{\Bdeen}[1]{%
\pgfmathsetmacro{\Min}{26.62}%
\pgfmathsetmacro{\Max}{30.14}%
\pgfmathsetmacro{\PercentColor}{max(min(100.0 * (#1 - \Min)/(\Max - \Min), 100.0), 0.00)}%
\cellScaleColor{\PercentColor}{#1}
}

\newcommand{\Cdeen}[1]{%
\pgfmathsetmacro{\Min}{.5433}%
\pgfmathsetmacro{\Max}{.5732}%
\pgfmathsetmacro{\PercentColor}{max(min(100.0 * (#1 - \Min)/(\Max - \Min), 100.0), 0.00)}%
\cellScaleColor{\PercentColor}{#1}
}

\newcommand{\Mdeen}[1]{%
\pgfmathsetmacro{\Min}{35.07}%
\pgfmathsetmacro{\Max}{37.42}%
\pgfmathsetmacro{\PercentColor}{max(min(100.0 * (#1 - \Min)/(\Max - \Min), 100.0), 0.00)}%
\cellScaleColor{\PercentColor}{#1}
}

\newcommand{\Ndeen}[1]{%
\pgfmathsetmacro{\Min}{-0.451}%
\pgfmathsetmacro{\Max}{-0.364}%
\pgfmathsetmacro{\PercentColor}{max(min(100.0 * (#1 - \Min)/(\Max - \Min), 100.0), 0.00)}%
\cellScaleColor{\PercentColor}{#1}
}

\newcommand{\Bencs}[1]{%
\pgfmathsetmacro{\Min}{18.24}%
\pgfmathsetmacro{\Max}{21.07}%
\pgfmathsetmacro{\PercentColor}{max(min(100.0 * (#1 - \Min)/(\Max - \Min), 100.0), 0.00)}%
\cellScaleColor{\PercentColor}{#1}
}

\newcommand{\Cencs}[1]{%
\pgfmathsetmacro{\Min}{0.4645}%
\pgfmathsetmacro{\Max}{0.4899}%
\pgfmathsetmacro{\PercentColor}{max(min(100.0 * (#1 - \Min)/(\Max - \Min), 100.0), 0.00)}%
\cellScaleColor{\PercentColor}{#1}
}

\newcommand{\Mencs}[1]{%
\pgfmathsetmacro{\Min}{24.57}%
\pgfmathsetmacro{\Max}{25.98}%
\pgfmathsetmacro{\PercentColor}{max(min(100.0 * (#1 - \Min)/(\Max - \Min), 100.0), 0.00)}%
\cellScaleColor{\PercentColor}{#1}
}

\newcommand{\Nencs}[1]{%
\pgfmathsetmacro{\Min}{-1.032}%
\pgfmathsetmacro{\Max}{-0.816}%
\pgfmathsetmacro{\PercentColor}{max(min(100.0 * (#1 - \Min)/(\Max - \Min), 100.0), 0.00)}%
\cellScaleColor{\PercentColor}{#1}
}

\newcommand{\Rencs}[1]{%
\pgfmathsetmacro{\Min}{81.28}%
\pgfmathsetmacro{\Max}{84.87}%
\pgfmathsetmacro{\PercentColor}{max(min(100.0 * (#1 - \Min)/(\Max - \Min), 100.0), 0.00)}%
\cellScaleColor{\PercentColor}{#1}
}

\newcommand{\Tencs}[1]{%
\pgfmathsetmacro{\Min}{4.48}%
\pgfmathsetmacro{\Max}{5.84}%
\pgfmathsetmacro{\PercentColor}{max(min(100.0 * (#1 - \Min)/(\Max - \Min), 100.0), 0.00)}%
\cellScaleColor{\PercentColor}{#1}
}

\newcommand{\Bentr}[1]{%
\pgfmathsetmacro{\Min}{11.55}%
\pgfmathsetmacro{\Max}{13.11}%
\pgfmathsetmacro{\PercentColor}{max(min(100.0 * (#1 - \Min)/(\Max - \Min), 100.0), 0.00)}%
\cellScaleColor{\PercentColor}{#1}
}

\newcommand{\Centr}[1]{%
\pgfmathsetmacro{\Min}{0.4500}%
\pgfmathsetmacro{\Max}{0.4620}%
\pgfmathsetmacro{\PercentColor}{max(min(100.0 * (#1 - \Min)/(\Max - \Min), 100.0), 0.00)}%
\cellScaleColor{\PercentColor}{#1}
}

\newcommand{\Nentr}[1]{%
\pgfmathsetmacro{\Min}{-0.916}%
\pgfmathsetmacro{\Max}{-0.657}%
\pgfmathsetmacro{\PercentColor}{max(min(100.0 * (#1 - \Min)/(\Max - \Min), 100.0), 0.00)}%
\cellScaleColor{\PercentColor}{#1}
}

\newcommand{\Gr}[1]{%
\cellScaleColor{0}{\color{Gray}#1}
}

\adjustbox{max width=\textwidth}{%
\begin{tabular}{ll|cccccccc|ccc}
\toprule

& & \multicolumn{8}{c|}{From random initialization}
& \multicolumn{3}{c|}{Direct finetuning from} \\

\cmidrule(lr){3-10} \cmidrule(lr){11-13} 

& & 32k & 16k & 8k & 4k & 2k & 1k & 500 & 0 &
500 & 1k & 2k
\\ \midrule


\multirow{6}{*}{\rotatebox{90}{en-de}}
& \multirow{2}{*}{BLEU} &
\Gr{26.9} & \Gr{26.9} & \Gr{26.7} & \Gr{26.4} & \Gr{26.4} & \Gr{26.1} & \Gr{25.8} & \multirow{2}{*}{\Bende{22.6}} &
    \Gr{25.2} & \Gr{25.0} & \Gr{25.0}
\\
    & & 
		\Bende{25.7} & \Bende{26.3} & \Bende{25.9} & \Bende{26.2} & \Bende{25.6} & \Bende{25.7} & \Bende{25.3} & &
    \Bende{24.9} & \Bende{24.3} & \Bende{24.7}

\\ \cdashlinelr{2-13}

& \multirow{2}{*}{chrF} &
		\Gr{.569} & \Gr{.568} & \Gr{.568} & \Gr{.568} & \Gr{.564} & \Gr{.564} & \Gr{.561} & \multirow{2}{*}{\Bende{.526}} &
    \Gr{.559} & \Gr{.559} & \Gr{.559}
\\
		& & 
		\Cende{.563} & \Cende{.565} & \Cende{.565} & \Cende{.568} & \Cende{.561} & \Cende{.561} & \Cende{.559} & &
    \Cende{.560} & \Cende{.553} & \Cende{.557}

\\ \cdashlinelr{2-13}

& \multirow{2}{*}{METEOR} &
		\Gr{47.7} & \Gr{48.0} & \Gr{47.9} & \Gr{47.8} & \Gr{47.9} & \Gr{47.7} & \Gr{47.6} & \multirow{2}{*}{\Mende{45.0}} &
    \Gr{46.5} & \Gr{46.4} & \Gr{46.4}
	\\
	& & 
	\Mende{47.0} & \Mende{47.8} & \Mende{47.4} & \Mende{48.0} & \Mende{47.5} & \Mende{47.8} & \Mende{47.7} & &
 \Mende{46.5} & \Mende{46.1} & \Mende{46.3}

\\ \midrule


\multirow{6}{*}{\rotatebox{90}{de-en}}
& \multirow{2}{*}{BLEU} &
	\Gr{29.8} & \Gr{30.1} & \Gr{29.6} & \Gr{29.3} & \Gr{28.6} & \Gr{28.5} & \Gr{28.1} & \multirow{2}{*}{\Bdeen{26.6}} &
    \Gr{28.2} & \Gr{28.4} & \Gr{27.7}
\\
& &
\Bdeen{29.8} & \Bdeen{29.3} & \Bdeen{28.8} & \Bdeen{29.5} & \Bdeen{28.7} & \Bdeen{28.8} & \Bdeen{28.6} & &
    \Bdeen{28.5} & \Bdeen{27.9} & \Bdeen{28.5}

\\ \cdashlinelr{2-13}

& \multirow{2}{*}{chrF} &
	\Gr{.570} & \Gr{.573} & \Gr{.568} & \Gr{.567} & \Gr{.562} & \Gr{.558} & \Gr{.558} & \multirow{2}{*}{\Cdeen{.543}} &
    \Gr{.562} & \Gr{.564} & \Gr{.559}
\\
& &
\Cdeen{.573} & \Cdeen{.570} & \Cdeen{.569} & \Cdeen{.571} & \Cdeen{.565} & \Cdeen{.566} & \Cdeen{.566} & &
\Cdeen{.564} & \Cdeen{.561} & \Cdeen{.565}

\\ \cdashlinelr{2-13}

& \multirow{2}{*}{METEOR} &
	\Gr{37.1} & \Gr{37.4} & \Gr{37.2} & \Gr{37.2} & \Gr{36.9} & \Gr{37.2} & \Gr{36.9} & \multirow{2}{*}{\Mdeen{35.1}} &
    \Gr{36.4} & \Gr{36.4} & \Gr{36.0} \\
& &
\Mdeen{37.0} & \Mdeen{37.1} & \Mdeen{36.9} & \Mdeen{37.2} & \Mdeen{37.0} & \Mdeen{37.0} & \Mdeen{37.0} & &
\Mdeen{36.5} & \Mdeen{36.3} & \Mdeen{36.5}

\\ \midrule


\multirow{6}{*}{\rotatebox{90}{en-cs}}
& \multirow{2}{*}{BLEU} &
	\Gr{21.1} & \Gr{20.8} & \Gr{20.9} & \Gr{20.6} & \Gr{20.1} & \Gr{20.0} & \Gr{19.5} & \multirow{2}{*}{\Bencs{18.2}} &
    \Gr{19.2} & \Gr{19.3} & \Gr{19.4}
\\
& &
\Bencs{20.7} & \Bencs{20.7} & \Bencs{20.7} & \Bencs{20.3} & \Bencs{20.0} & \Bencs{20.0} & \Bencs{19.7} & &
\Bencs{19.5} & \Bencs{19.0} & \Bencs{19.7}

\\ \cdashlinelr{2-13}

& \multirow{2}{*}{chrF} &
	\Gr{.489} & \Gr{.488} & \Gr{.490} & \Gr{.487} & \Gr{.483} & \Gr{.482} & \Gr{.478} & \multirow{2}{*}{\Cencs{.465}} &
    \Gr{.477} & \Gr{.476} & \Gr{.478}
\\
& &
\Cencs{.488} & \Cencs{.489} & \Cencs{.488} & \Cencs{.486} & \Cencs{.484} & \Cencs{.482} & \Cencs{.480} & &
\Cencs{.480} & \Cencs{.475} & \Cencs{.482}

\\ \cdashlinelr{2-13}

& \multirow{2}{*}{METEOR} &
	\Gr{26.0} & \Gr{25.8} & \Gr{26.0} & \Gr{25.8} & \Gr{25.7} & \Gr{25.7} & \Gr{25.4} & \multirow{2}{*}{\Mencs{24.6}} &
    \Gr{25.2} & \Gr{25.2} & \Gr{25.2} \\
& &
\Mencs{25.7} & \Mencs{25.8} & \Mencs{25.9} & \Mencs{25.7} & \Mencs{25.6} & \Mencs{25.7} & \Mencs{25.7} & &
\Mencs{25.1} & \Mencs{24.8} & \Mencs{25.1}

\\ \midrule


\multirow{4}{*}{\rotatebox{90}{en-tr}}
& \multirow{2}{*}{BLEU} &
	\Gr{12.6} & \Gr{13.1} & \Gr{12.7} & \Gr{12.8} & \Gr{12.5} & \Gr{12.3} & \Gr{12.2} & \multirow{2}{*}{\Bentr{12.4}} &
    \Gr{12.0} & \Gr{12.6} & \Gr{12.3}
\\
& &
\Bentr{10.7} & \Bentr{11.6} & \Bentr{12.2} & \Bentr{12.7} & \Bentr{12.6} & \Bentr{12.5} & \Bentr{12.5} & &
\Bentr{12.3} & \Bentr{12.2} & \Bentr{12.6}

\\ \cdashlinelr{2-13}
& \multirow{2}{*}{chrF} &
	\Gr{.455} & \Gr{.462} & \Gr{.459} & \Gr{.456} & \Gr{.457} & \Gr{.457} & \Gr{.455} & \multirow{2}{*}{\Centr{.461}} &
    \Gr{.456} & \Gr{.460} & \Gr{.459}

	\\

& & \Centr{.436} & \Centr{.446} & \Centr{.457} & \Centr{.461} & \Centr{.464} & \Centr{.461} & \Centr{.459} & &
\Centr{.460} & \Centr{.461} & \Centr{.464}

\\ \bottomrule

\end{tabular}}

%% file: validation_table.tex
\adjustbox{max width=\textwidth}{%
\begin{tabular}{ll|cccccccc|ccc|c}
\toprule

& & \multicolumn{8}{c|}{From random initialization}
& \multicolumn{3}{c|}{Direct finetuning from}
& \multirow{2}{*}{In steps} \\

\cmidrule(lr){3-10} \cmidrule(lr){11-13} 

& & 32k & 16k & 8k & 4k & 2k & 1k & 500 & 0 &
500 & 1k & 2k & 
\\ \midrule


en-de
& &
    29.07 & 29.76 & 28.6 & 28.7 & 28.11 & 27.61 & 27.66 & 26.09 &
    28.04 & 27.89 & 27.87 &
    27.75

\\ \midrule


de-en &  &
    35.05 & 35.26 & 34.34 & 35.34 & 34.37 & 34.84 & 33.83 & 27.96 &
    32.61 & 33.47 & 33.68 &
    32.44
\\ \midrule


en-cs & &
    22.47 & 22.45 & 22.53 & 22.29 & 21.94 & 21.78 & 21.49 & 20.26 &
    22.03 & 21.31 & 21.4 &
    21.14
\\ \midrule


en-tr & &
    13.40 & 14.18 & 14.25 & 14.11 & 14.05 & 13.72 & 13.94 & 14.55 &
    12.02 & 12.25 & 12.28 &
    11.56
\\ \bottomrule

\end{tabular}}